\documentclass[letterpaper]{article} 
\usepackage[draft]{aaai2026}  
\usepackage{times}  
\usepackage{helvet}  
\usepackage{courier}  
\usepackage[hyphens]{url}  
\usepackage{graphicx} 
\usepackage{natbib}  
\usepackage{caption} 
\usepackage{algorithm}
\usepackage{booktabs}
\usepackage{multirow}
\usepackage[most]{tcolorbox}
\usepackage{enumitem}
\tcbuselibrary{listings}

\usepackage{amsmath}
\usepackage{amssymb}
\usepackage{algpseudocode}

\tcbset{
  myexamplebox/.style={
    enhanced,
    colback=white,
    colframe=black!60,
    sharp corners,
    boxrule=0.5pt,
    fonttitle=\bfseries,
    title=#1,
    coltitle=black,
    top=2mm, bottom=2mm, left=2mm, right=2mm,
    width=\linewidth,
  }
}

\usepackage{newfloat}
\usepackage{listings}
\DeclareCaptionStyle{ruled}{labelfont=normalfont,labelsep=colon,strut=off} 
\floatstyle{ruled}
\newfloat{listing}{tb}{lst}{}
\floatname{listing}{Listing}
\title{GUI-Eyes: Tool-Augmented Perception for Visual Grounding in GUI Agents}
\author{
    Chen Chen\textsuperscript{\rm 1,3}, 
    Jiawei Shao\textsuperscript{\rm 2}\thanks{Corresponding author.}, 
    Dakuan Lu\textsuperscript{\rm 2}, 
    Haoyi Hu\textsuperscript{\rm 4},\\ 
    Xiangcheng Liu\textsuperscript{\rm 1,3}, 
    Hantao Yao\textsuperscript{\rm 1}, 
    Wu Liu\textsuperscript{\rm 1}
}
\affiliations{
    \textsuperscript{\rm 1}University of Science and Technology of China\\
    \textsuperscript{\rm 2}Institute of Artificial Intelligence (TeleAI), China Telecom\\
    \textsuperscript{\rm 3}Shanghai Innovation Institute\\
    \textsuperscript{\rm 4}Shanghai Jiao Tong University\\
    chenchen0318@mail.ustc.edu.cn
}

\usepackage{bibentry}

\begin{document}

\maketitle

\begin{abstract}
Recent advances in vision-language models (VLMs) and reinforcement learning (RL) have driven progress in GUI automation. However, most existing methods rely on static, one-shot visual inputs and passive perception, lacking the ability to adaptively determine when, whether, and how to observe the interface.
We present GUI-Eyes, a reinforcement learning framework for active visual perception in GUI tasks. To acquire more informative observations, the agent learns to make strategic decisions on both whether and how to invoke visual tools, such as cropping or zooming, within a two-stage reasoning process.
To support this behavior, we introduce a progressive perception strategy that decomposes the decision-making into coarse exploration and fine-grained grounding, coordinated by a two-level policy. In addition, we design a spatially continuous reward function tailored to tool usage, which integrates both location proximity and region overlap to provide dense supervision and alleviate the reward sparsity common in GUI environments.
On the ScreenSpot-Pro benchmark, GUI-Eyes-3B achieves 44.8\% grounding accuracy using only 3k labeled samples, significantly outperforming both supervised and RL-based baselines. These results highlight that tool-aware active perception, enabled by staged policy reasoning and fine-grained reward feedback, is critical for building robust and data-efficient GUI agents.

\end{abstract}


\section{Introduction}

The development of large language models (LLMs)~\cite{llama2,llama3,qwen2,qwen3} and vision language models (VLMs)~\cite{qwen2-vl,qwen2.5-vl,gpt4,gpt4O} has introduced new opportunities and challenges in applying these models to GUI tasks.
Existing GUI agents are mainly based on supervised fine-tuning (SFT)~\cite{os-atlas,ui-tars,cogagent}, where large annotated datasets are used to teach model interface understanding and action planning. However, this approach suffers from several limitations: it requires expensive, labor-intensive data collection, and the resulting models often lack robustness when deployed in unfamiliar or out-of-domain environments~\cite{amex,simple_test-time}.

DeepSeek-R1~\cite{deepseek-r1} demonstrates that reinforcement learning (RL) can enhance large models’ problem-solving ability without human-labeled data, by optimizing behavior through interaction and well-designed rewards. In GUI tasks, RL effectively reduces supervision demands while improving adaptability and generalization~\cite{gui-r1,infigui-r1,ui-r1,gui—g1,uishift}, making it a promising alternative to SFT. Despite this potential, most existing methods still optimize only textual outputs~\cite{ui-r1,gui-r1,infigui-r1,enhancing}, overlooking visual cues essential for GUI understanding. Real users rely on visual attention to locate elements and interpret layouts~\cite{deepeyes,visual_planning,cogagent,pixel}; models limited to language reasoning struggle with ambiguous instructions and complex interfaces. Thus, GUI agents must integrate perception and decision-making—learning not only what to see but how often to observe—to support robust visual-grounded interactions.

To address the limitations of supervised learning in terms of data annotation cost and generalization, we propose \textbf{GUI-Eyes}, a reinforcement learning framework centered on active perception. Unlike traditional GUI agents that rely on static, one-shot visual input, GUI-Eyes empowers the model to dynamically decide whether to invoke visual tools during reasoning, and to flexibly configure their parameters (e.g., crop region, zoom scale) for acquiring task-relevant observations step by step. By modeling visual perception as an optimizable policy, GUI-Eyes learns a \textbf{perception--reasoning--perception }loop that tightly coordinates visual observation with language-based decision-making. 

\begin{figure}[t]
\centering
\includegraphics[width=\linewidth]{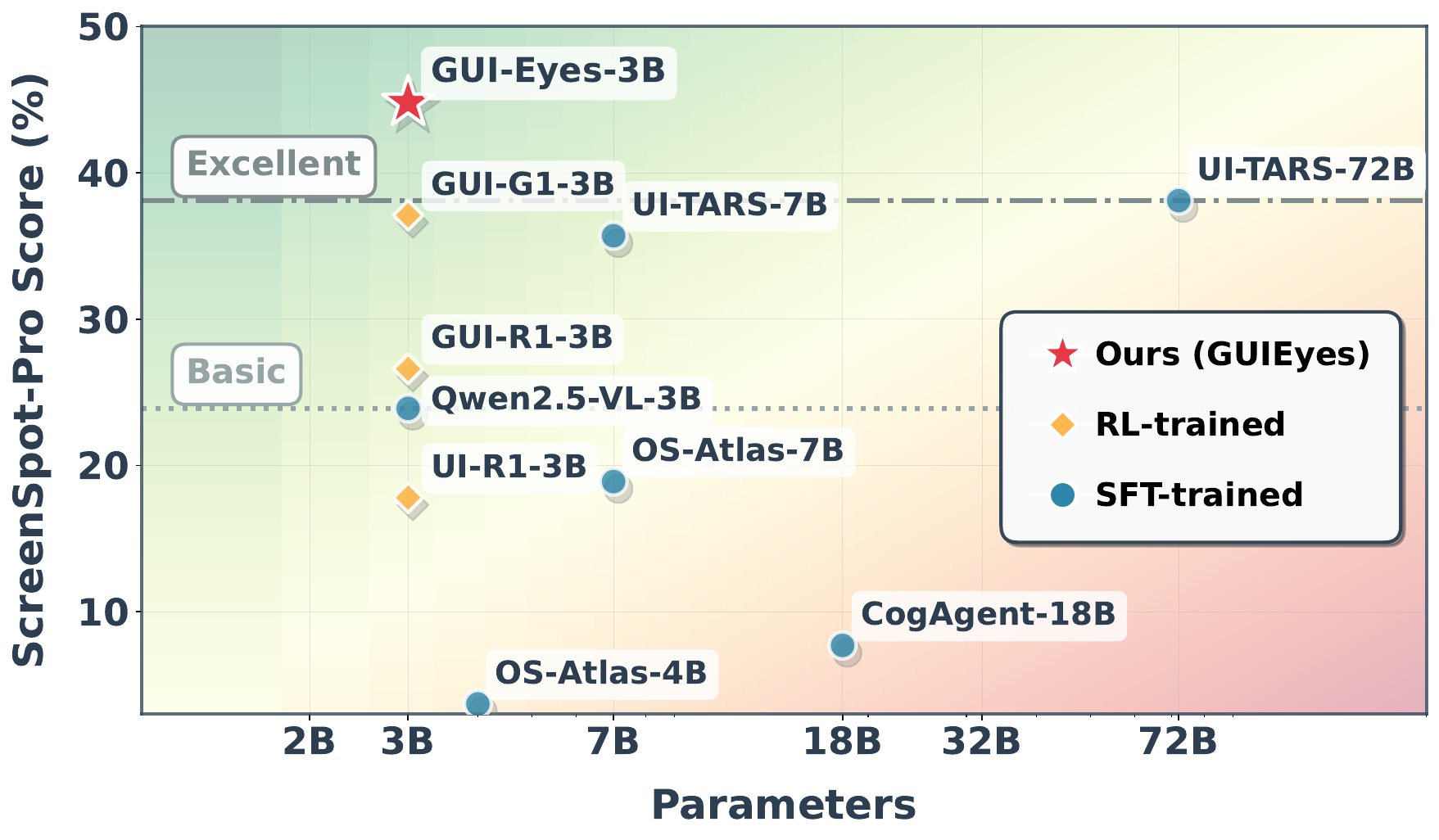}
\caption{Performance Scaling of Multimodal UI Understanding Models on the ScreenSpot-Pro Benchmark. Our method achieves state-of-the-art performance.}
\label{fig:performance_scaling}
\end{figure}

As shown in Figure~\ref{fig:performance_scaling}, GUI-Eyes-3B achieves superior performance on the ScreenSpot-Pro benchmark compared to existing models of similar or larger scales, validating the effectiveness of our tool-aware active perception framework.

\noindent\textbf{Our main contributions are summarized as follows:}
\begin{itemize}
    \item We propose \textbf{GUI-Eyes}, a novel framework that integrates active perception into GUI agents. It enables the model to autonomously determine \emph{when} and \emph{how} to invoke visual tools during reasoning, achieving more precise and task-adaptive visual understanding.

    \item To support the learning of effective active perception behaviors, we design a multi-factor reward function that provides structured supervision over format correctness, initial localization, and spatial coverage. This facilitates more stable and generalizable tool-use policy learning.

    \item We conduct extensive experiments on the ScreenSpot-Pro benchmark. GUI-Eyes-3B achieves \textbf{44.8\%} grounding accuracy using only \textbf{3,000 labeled samples}, significantly outperforming both supervised and RL-based baselines, thereby demonstrating strong sample efficiency and robust generalization.
\end{itemize}

\section{Related Work}

\subsection{GUI Agents}

With the growing capabilities of multimodal large language models (MLLMs)~\cite{gpt4,qwen2-vl,qwen2.5-vl,gpt4O}, GUI-based agents have become a central focus in human–computer interaction research~\cite{mobileagentv1,mobileagentv2,gui_survey,gui_survey2}. Existing work in this area can be broadly categorized into two main paradigms: structure-driven and vision-driven approaches.

\noindent\textbf{Structure-driven methods} rely on structured representations such as HTML or DOM trees to parse and execute interface instructions~\cite{realworld_agent,language_solve,mind2web,webarena,androidinthewild}. These methods benefit from explicit symbolic semantics and direct access to internal interface states.

\noindent\textbf{Vision-driven methods}, in contrast, operate directly on GUI screenshots, leveraging visual perception and language instructions for open-ended reasoning and grounding~\cite{appagent,cogagent,ferret,infiguiagent,gui-actor}. For instance, AppAgent~\cite{appagent} explores autonomous interactions in mobile applications, while CogAgent~\cite{cogagent} uses high-resolution visual encoders for better UI understanding. OS-Atlas~\cite{os-atlas} proposes a unified action representation and pre-trains on 13M UI elements to enable cross-platform generalization. ScaleTrack~\cite{scaletrack} designs a backtracking-based task to enhance multi-step reasoning, training on 7.5M screenshots for joint grounding and planning.
Despite strong empirical results, these methods mainly depend on supervised learning with static input–output pairs, limiting their ability to actively acquire perceptual information or adjust reasoning strategies during inference.

\begin{figure*}[t]
\centering
\includegraphics[width=0.98\linewidth]{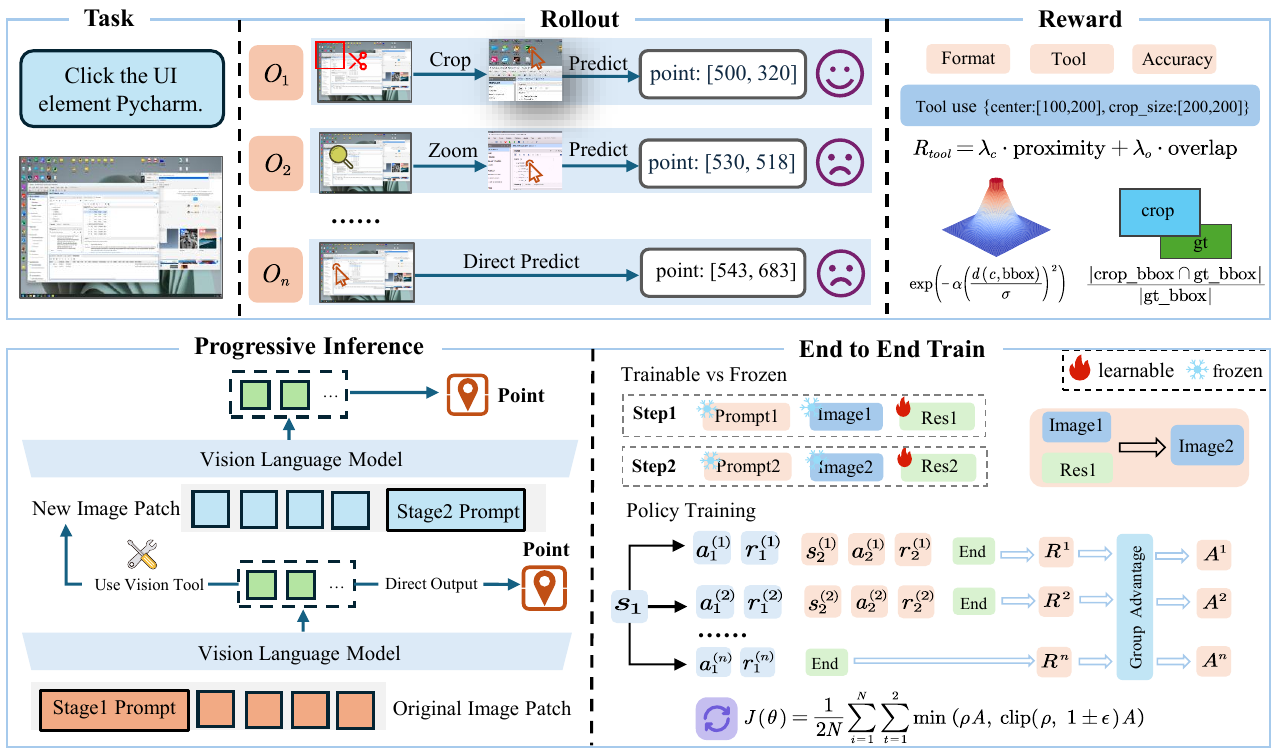}
\caption{
Overview of the GUI-Eyes Framework. The top illustrates a rollout example with optional visual tool invocation, together with a tool-specific reward function that combines spatial proximity and region overlap relative to the ground-truth. The bottom depicts the progressive inference architecture and end-to-end training pipeline, where the two-stage decision process is guided by stage-specific prompts, and visual inputs are dynamically generated through previously applied visual tools.
}

\label{fig:pipeline}
\end{figure*}

\begin{figure}[t]
\centering
\includegraphics[width=\linewidth]{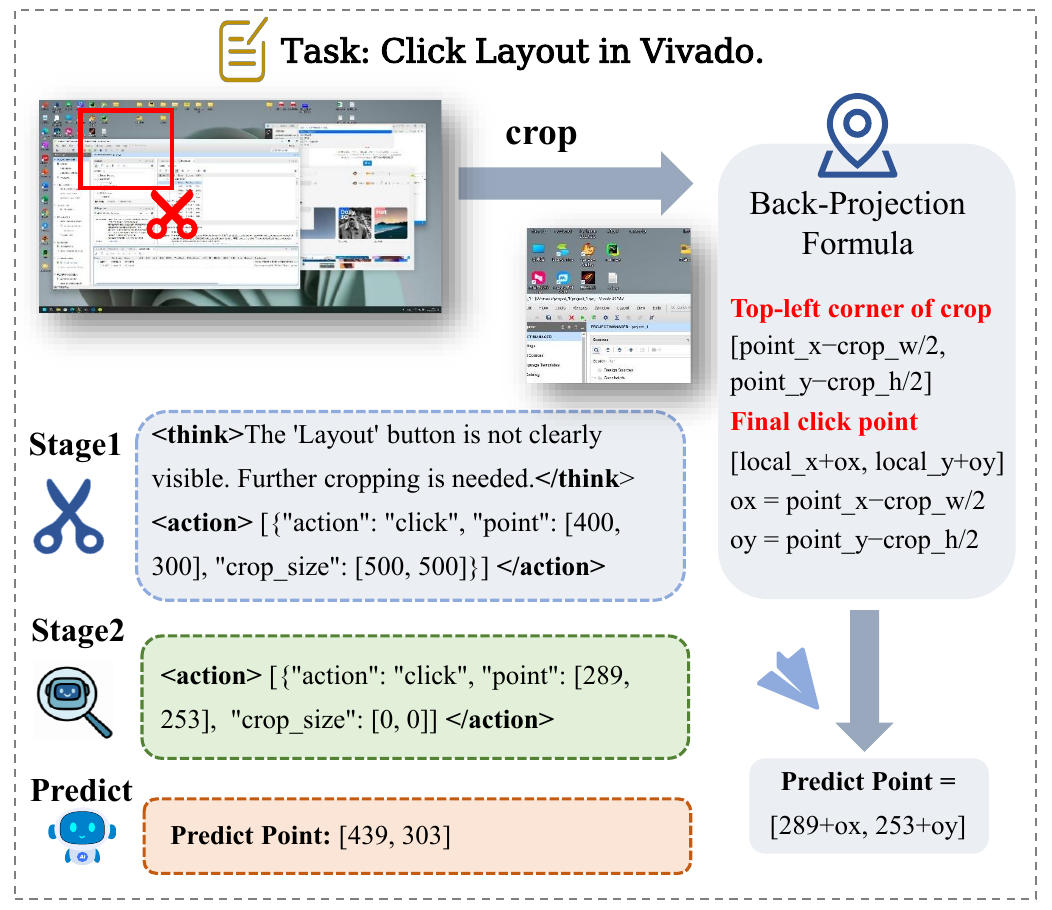}
\caption{Inference Example of Tool-Augmented Reasoning with Cropping in a GUI Task.}
\label{fig:example-inference}
\end{figure}

\subsection{Reinforcement Fine-Tuning}

Traditional supervised fine-tuning (SFT) for GUI agents requires extensive annotated data~\cite{os-atlas,ui-tars,scaletrack,mobilevlm}, making it costly and less generalizable. Recently, rule-based reinforcement learning (RL) has emerged as a more efficient alternative, especially in low-resource scenarios. DeepSeek-R1~\cite{deepseek-r1} introduced this paradigm for large language models by employing predefined reward functions (e.g., symbolic correctness) to evaluate outputs without human feedback. This idea has been extended to GUI agents~\cite{ui-r1,gui-r1,gui—g1,infigui-r1,uishift,web-r1,mobile-r1}, showing strong data efficiency.

UI-R1~\cite{ui-r1} first introduced rule-based RL for low-level GUI action prediction, achieving strong performance with only 130 mobile samples and substantially lowering data requirements. InfiGUI-R1~\cite{infigui-r1} proposed the Actor2Reasoner framework, encouraging agents to “think before acting and reflect after execution,” thereby improving their understanding and operation in complex UI layouts. Interestingly, GUI-G1~\cite{gui—g1} showed that, for some tasks, directly producing the final answer can even surpass step-wise reasoning, indicating that the utility of intermediate reasoning varies with task complexity and type.

Although these methods have achieved notable progress in data efficiency and policy learning, they still rely mainly on text-only reasoning~\cite{vagen,ragen,search-r1}, overlooking the role of visual information in complex GUI environments. Inspired by advances in visual reasoning and active perception~\cite{deepeyes,visualtoolagent,visual_planning}, we argue that GUI agents should proactively observe and reason—leveraging visual tools to interpret screenshots and understand task-specific visual contexts.

This integration enhances situational awareness and boosts performance in visually complex or ambiguous GUI scenarios, following recent studies emphasizing collaborative reasoning between language and vision modules~\cite{aiflow1,aiflow2}.

\section{GUI-Eyes}

In this section, we introduce our method, including the Progressive Inference (3.1), Reward Design (3.2), and the Training Details (3.3). An overview of the overall architecture is shown in Figure~\ref{fig:pipeline}.

\subsection{Progressive Inference}
Recent GUI agents typically rely on static, one-shot visual input and pre-defined tool usage, lacking the ability to actively perceive task-relevant information during reasoning. 

To overcome this limitation, we propose \textbf{GUI-Eyes}, a reinforcement learning framework that empowers agents with active perception capabilities. Instead of relying on fixed visual inputs, GUI-Eyes learns to strategically decide when and how to observe the GUI environment via a set of visual tools (e.g., cropping, zooming). The model forms a perception--reasoning--perception loop, enabling dynamic attention and adaptive visual understanding.

\noindent\textbf{Stage 1: Active Perception Planning.} \\
Given a natural language instruction and the original GUI screenshot, the model performs an initial grounding attempt. It then autonomously decides whether to invoke a visual tool and predicts its configuration parameters, such as the crop center, region size, and zoom scale. These parameters are used to generate an intermediate visual input (e.g., a cropped or zoomed image), which serves as refined perceptual input for the next stage.

As illustrated in the top part of Figure~\ref{fig:pipeline}, the agent may choose different rollouts, such as applying Crop, Zoom, or directly predicting without a tool.

\noindent\textbf{Stage 2: Reasoning with Focused Perception.} \\
In the second stage, the model conducts more fine-grained reasoning based on the intermediate input. By focusing on visually clearer and task-relevant regions, the model enhances the accuracy and robustness of its prediction, particularly under high-resolution, cluttered, or ambiguous interface conditions.

This two-stage process enables the agent to interactively refine its perception based on task-specific feedback. A representative example of active visual grounding, in which the model invokes the Crop tool, is shown in Figure~\ref{fig:example-inference}.

Rather than concatenating multi-turn inputs, we treat each inference round as a perception-informed decision point. The output of the previous step—such as the chosen visual tool and its manipulated image—reshapes the visual input for the next.
As illustrated in the lower-left part of Figure~\ref{fig:pipeline}, this process enables the agent to progressively refine its focus across reasoning steps, forming a closed-loop cycle of perception, reasoning, and re-perception.

\subsection{Reward Design for Reinforcement Learning}

As illustrated in the bottom-right part of Figure~\ref{fig:pipeline}, 
GUI-Eyes performs end-to-end reinforcement learning that jointly optimizes perception and reasoning policies. 
Each training trajectory consists of two decision steps—Stage 1 for visual tool planning and Stage 2 for task execution—enabling a unified optimization of the progressive inference process.

To guide this optimization, we design a unified reward function that integrates perception-aware actions and outcome-level accuracy into a single learning signal:

\begin{equation}
R(\tau) = \lambda_{\text{acc}} R_{\text{acc}} + \lambda_{\text{format}} R_{\text{format}} + \lambda_{\text{tool}} R_{\text{tool}}
\label{eq:total_reward}
\end{equation}

\noindent\textbf{Format Reward} $R_{\text{format}}$:  
Encourages syntactic correctness by penalizing malformed tags or invalid actions.

\noindent\textbf{Accuracy Reward} $R_{\text{acc}}$:
A binary reward that assigns 1 if the predicted point falls within the ground-truth bounding box, and 0 otherwise.

\noindent\textbf{Tool Reward} $R_{\text{tool}}$:  
Reflects the quality of tool usage, combining the proximity of the selected center \textbf{$c$} to the target and the region coverage:

\begin{equation}
\begin{aligned}
R_{\text{tool}} =\;& \lambda_{\text{center}} \cdot 
\exp\left( -\alpha \left( \frac{d(c, \text{gt\_bbox})}{\sigma} \right)^2 \right) \\
&+ \lambda_{\text{overlap}} \cdot 
\frac{|\text{crop\_bbox} \cap \text{gt\_bbox}|}{|\text{gt\_bbox}|}
\end{aligned}
\label{eq:tool_reward}
\end{equation}

Here, $\text{gt\_bbox}$ denotes the ground-truth bounding box of the target element, and $\text{crop\_bbox}$ represents the region produced by the model's tool action.  
The function $d(c, \text{gt\_bbox})$ computes the shortest distance from the selected center $c$ to the boundary of $\text{gt\_bbox}$.  
The weighting factors $\lambda_{\text{center}}$ and $\lambda_{\text{overlap}}$ balance the contributions of the center alignment and spatial coverage terms.

\textit{Remark:} Our framework supports both \textbf{Crop} and \textbf{Zoom} tools, which share the same spatial parameters (center and size). Since zooming can be viewed as a visual transformation of cropping, we apply the same reward function to both, allowing for unified training and supervision. 

When the model decides not to invoke any visual tool in the first stage, we set $\text{crop\_size}=[0, 0]$. In this case, the IoU term of $R_{\text{tool}}$ becomes zero, but the center term still gives a small positive reward if the predicted location is close to the ground-truth region. This encourages the model to make accurate direct predictions for simpler tasks, preventing unnecessary tool usage and promoting adaptive tool invocation based on task complexity.

\begin{table*}[t]
\centering
\small
\setlength{\tabcolsep}{5.5pt}
\begin{tabular}{lccccccccccccccc}
\toprule
\multirow{2}{*}{\textbf{Model}} & 
\multicolumn{2}{c}{\textbf{CAD}} & 
\multicolumn{2}{c}{\textbf{Development}} & 
\multicolumn{2}{c}{\textbf{Creative}} & 
\multicolumn{2}{c}{\textbf{Scientific}} & 
\multicolumn{2}{c}{\textbf{Office}} & 
\multicolumn{2}{c}{\textbf{OS}} & 
\multicolumn{3}{c}{\textbf{Avg.}} \\
\cmidrule(lr){2-3}   
\cmidrule(lr){4-5} 
\cmidrule(lr){6-7}
\cmidrule(lr){8-9}
\cmidrule(lr){10-11}
\cmidrule(lr){12-13}
\cmidrule(lr){14-16}
 & Text & Icon & 
   Text & Icon & 
   Text & Icon & 
   Text & Icon & 
   Text & Icon & 
   Text & Icon & 
   Text & Icon & Avg \\
\midrule
\multicolumn{16}{l}{\textbf{Proprietary Models}} \\
GPT-4o & 2.0 & 0.0 & 1.3 & 0.0 & 1.0 & 0.0 & 2.1 & 0.0 & 1.1 & 0.0 & 0.0 & 0.0 & 1.3 & 0.0 & 0.8 \\
Claude Computer Use & 14.5 & 3.7 & 22.0 & 3.9 & 25.9 & 3.4 & 33.9 & 15.8 & 30.1 & 16.3 & 11.0 & 4.5 & 23.4 & 7.1 & 17.1 \\
\midrule
\multicolumn{16}{l}{\textbf{General Open-source Models}} \\
Qwen2.5-VL-3B & 9.1 & 7.3 & 22.1 & 1.4 & 26.8 & 2.1 & 38.2 & 7.3 & 33.9 & 15.1 & 10.3 & 1.1 & 23.6 & 3.8 & 16.1 \\
Qwen2.5-VL-7B & 16.8 & 1.6 & 46.8 & 4.1 & 35.9 & 7.7 & 49.3 & 7.3 & 52.5 & 20.8 & 37.4 & 6.7 & 38.9 & 7.1 & 26.8 \\
\midrule
\multicolumn{16}{l}{\textbf{GUI-Specific Models(SFT(+RL))}} \\
CogAgent-18B & 7.1 & 3.1 & 14.9 & 0.7 & 9.6 & 0.0 & 22.2 & 1.8 & 13.0 & 0.0 & 5.6 & 0.0 & 12.0 & 0.8 & 7.7 \\
OS-Atlas-7B & 12.2 & 4.7 & 33.1 & 1.4 & 28.8 & 2.8 & 37.5 & 7.3 & 33.9 & 5.7 & 27.1 & 4.5 & 28.1 & 4.0 & 18.9 \\
ShowUI-2B & 2.5 & 0.0 & 16.9 & 1.4 & 9.1 & 0.0 & 13.2 & 7.3 & 15.3 & 7.5 & 10.3 & 2.2 & 10.8 & 2.6 & 7.7 \\
UGround-7B & 14.2 & 1.6 & 26.6 & 2.1 & 27.3 & 2.8 & 31.9 & 2.7 & 31.6 & 11.3 & 17.8 & 0.0 & 25.0 & 2.8 & 16.5 \\
UGround-V1-7B & 15.8 & 1.2 & 51.9 & 2.8 & 47.5 & 9.7 & 57.6 & 14.5 & 60.5 & 13.2 & 38.3 & 7.9 & 45.2 & 8.1 & 31.1 \\
UI-TARS-2B & 17.8 & 4.7 & 47.4 & 4.1 & 42.9 & 6.3 & 56.9 & 17.3 & 50.3 & 17.0 & 21.5 & 5.6 & 39.6 & 8.4 & 27.7 \\
UI-TARS-7B & 20.8 & 9.4 & 58.4 & 12.4 & 50.0 & 9.1 & 63.9 & \textbf{31.8} & 63.3 & 20.8 & 30.8 & 16.9 & 47.8 & 16.2 & 35.7 \\
InfiGUI-R1-3B & 33.0 & \textbf{14.1} & 51.3 & 12.4 & 44.9 & 7.0 & 58.3 & 20.0 & 65.5 & 28.3 & 43.9 & 12.4 & 49.1 & 14.1 & 35.7 \\
\midrule
\multicolumn{16}{l}{\textbf{GUI-Specific Models(RL Only)}} \\
UI-R1-3B & 11.2 & 6.3 & 22.7 & 4.1 & 27.3 & 3.5 & 42.4 & 11.8 & 32.2 & 11.3 & 13.1 & 4.5 & 24.9 & 6.4 & 17.8 \\
GUI-R1-3B & 26.4 & 7.8 & 33.8 & 4.8 & 40.9 & 5.6 & 61.8 & 17.3 & 53.6 & 17.0 & 28.1 & 5.6 & 45.1 & 8.1 & 30.2 \\
GUI-R1-7B & 23.9 & 6.3 & 49.4 & 4.8 & 38.9 & 8.4 & 55.6 & 11.8 & 58.7 & 26.4 & 42.1 & 16.9 & 45.9 & 11.2 & 32.4 \\
SE-GUI-3B & 38.1 & 12.5 & 55.8 & 7.6 & 47.0 & 4.9 & 61.8 & 16.4 & 59.9 & 24.5 & 40.2 & 12.4 & 50.4 & 11.8 & 35.9 \\
GUI-G1-3B & 39.6 & 9.4 & 50.7 & 10.3 & 36.6 & 11.9 & 61.8 & 30.0 & 67.2 & \textbf{32.1} & 23.5 & 10.6 & 49.5 & \textbf{16.8 }& 37.1 \\
\midrule
\textbf{3B(ours)} & \textbf{48.2} & 9.4 & \textbf{70.8} & \textbf{12.4} & \textbf{56.6} & \textbf{13.3} & \textbf{69.4} & 19.1 & \textbf{75.7} & 24.5 & \textbf{59.8} & \textbf{20.2} & \textbf{62.8} & 15.7 & \textbf{44.8} 
\\
\bottomrule
\end{tabular}
\caption{Performance comparison of different agent models across various task categories based on Text, Icon, and Average scores on ScreenSpot-Pro. Results marked in \textbf{bold} represent the best performance.}
\label{tab:screenspot-pro}
\end{table*}
\subsection{Policy Optimization and Training Details}

Building upon the unified reward described above, we optimize GUI-Eyes via end-to-end reinforcement learning. 
As illustrated in the bottom-right part of Figure~\ref{fig:pipeline}, the two-stage reasoning process (perception and decision) is trained jointly as a unified trajectory, where the reward signal consistently guides both visual perception and task execution.

\noindent\textbf{Advantage Computation.}  
We use the total reward $R(\tau)$ to guide policy learning. Following prior work (e.g., GRPO~\cite{grpo}), the advantage $A(a_t^{(i)})$ of each sampled response is computed by normalizing its reward within a batch. Specifically, given $N$ sampled responses $\{o_1, o_2, \dots, o_N\}$ with corresponding rewards $\{R_1, R_2, \dots, R_N\}$, the advantage for response $i$ is computed as:

\begin{align}
A_i = \frac{R_i - \text{mean}(R_1, R_2, ..., R_N)}{\text{std}(R_1, R_2, ..., R_N)}
\end{align}

\begin{table*}[t]
\centering
\small
\begin{tabular}{l c | ccc c | ccc c}
\toprule
\multirow{2}{*}{\textbf{Model}} & \multirow{2}{*}{\textbf{Training Samples}} & \multicolumn{4}{c|}{\textbf{ScreenSpot Accuracy (\%)}} & \multicolumn{4}{c}{\textbf{ScreenSpot-v2 Accuracy (\%)}} \\
& & Mobile & Desktop & Web & Avg. & Mobile & Desktop & Web & Avg. \\
\midrule
\multicolumn{10}{l}{\textbf{Proprietary Models}} \\
GPT-4o & - & 21.9 & 17.8 & 9.4 & 18.8 & 22.5 & 22.2 & 12.4 & 20.1 \\
\midrule
\multicolumn{10}{l}{\textbf{General Open-source Models}} \\
Qwen2-VL-7B & - & 50.3 & 40.4 & 27.4 & 42.9 & 39.4 & 50.1 & 27.7 & 39.8 \\
Qwen2.5-VL-3B & - & - & - & - & 55.5 & 55.5 & 44.0 & 39.1 & 46.9 \\
Qwen2.5-VL-7B & - & - & - & - & 84.7 & 92.8 & 78.4 & 85.4 & 86.5 \\
\midrule
\multicolumn{10}{l}{\textbf{GUI-Specific Models}} \\
CogAgent-18B & 222M & 57.8 & 31.6 & 40.1 & 47.4 & 50.6 & 51.6 & 54.1 & 52.8 \\
SeeClick-7B & 1M & 68.1 & 48.8 & 41.8 & 53.4 & 51.8 & 65.5 & 40.7 & 53.9 \\
UGround-7B & 10M & 75.9 & 75.8 & 78.3 & 73.3 & 74.3 & 74.9 & 78.6 & 76.3 \\
ShowUI-2B & 256K & 84.8 & 70.8 & 76.2 & 75.1 & 70.0 & 85.1 & 73.3 & 77.3 \\
OSAtlas-4B & 13M & 56.2 & 74.9 & 69.9 & 68.5 & 74.9 & 56.9 & 70.7 & 68.5 \\
OSAtlas-7B & 13M & 85.0 & 78.8 & 84.5 & 82.5 & 78.3 & 85.5 & 83.8 & 83.3 \\
Aguvis-7B & 1M & 86.9 & 82.4 & 84.7 & 84.4 & 89.6 & \textbf{86.8} & 84.9 & 87.3 \\
UI-TARS-2B & 2M & 85.0 & 81.4 & 79.8 & 82.3 & 87.9 & 81.4 & 82.9 & 84.7 \\
\midrule
\multicolumn{10}{l}{\textbf{Ours}} \\
\textbf{GUI-Eyes-3B} & 3K & \textbf{89.9} & \textbf{88.3} & \textbf{85.1} & \textbf{87.8} & \textbf{91.6} & 86.2 & \textbf{86.3} & \textbf{88.4} \\
\bottomrule
\end{tabular}
\caption{Comparison of model performance on ScreenSpot and ScreenSpot-v2. Results marked in \textbf{bold} represent the best performance.}
\label{tab:screenspot-full}
\end{table*}

We adopt an agent-centric variant of the GRPO algorithm~\cite{verl-agent} that supports multi-stage reasoning, treating each decision step as part of a unified trajectory. This enables joint optimization of visual perception and task execution policies via end-to-end reinforcement learning. Our optimization objective is formulated as follows:

\begin{align}
J(\theta)
&= \mathbb{E}_{x \sim p(X),\, \{\tau_i\} \sim \pi_{\theta_{\text{old}}}}
\Bigg[
\frac{1}{2N} \sum_{i=1}^{N} \sum_{t=1}^{2}
\min \Big( \nonumber\\
&\qquad
\rho_\theta(a_t^{(i)})A(a_t^{(i)}),\;
\operatorname{clip}(\rho_\theta(a_t^{(i)}), 1 \pm \epsilon)
A(a_t^{(i)})
\Big)
\Bigg]
\end{align}

Here, $\rho_\theta(a_t^{(i)}) = \frac{\pi_\theta(a_t^{(i)})}{\pi_{\theta_{\text{old}}}(a_t^{(i)})}$ is the importance sampling ratio between the current and old policies. $A(a_t^{(i)})$ denotes the estimated advantage of action $a_t^{(i)}$, computed based on normalized total rewards. The $\operatorname{clip}$ operator stabilizes updates by constraining the impact of large policy shifts.

Each sampled trajectory $\tau_i$ consists of two decision steps corresponding to our two-stage reasoning process ($t=1$ for perception, $t=2$ for final decision). The objective averages over $N$ samples and both reasoning steps per sample.

\section{Experiment}

In this section, we describe our experimental setup from three perspectives. 
Implementation Details outlines the training configuration, datasets, and evaluation benchmarks. 
The Experimental Results and Analysis section presents the performance of GUI-Eyes across various benchmarks, comparing it to state-of-the-art methods and offering further insights into task-specific behaviors.
Ablation Study analyzes the contribution of key components to overall performance.

\subsection{Implementation Details}

\textbf{Training Details.}  
We adopt Qwen2.5-VL-3B~\cite{qwen2.5-vl} as our base model and conduct training within the DeepEyes~\cite{deepeyes} framework using the GRPO algorithm~\cite{grpo}. Training is performed for \textbf{1} epoch with a batch size of \textbf{32} and a sampling temperature of $1.0$ to encourage exploration. Policy optimization is carried out using the AdamW optimizer~\cite{adamw} with a learning rate of \textbf{$1\times10^{-6}$}. All experiments are conducted on 8×NVIDIA H100-80G GPUs.

\noindent\textbf{Training Dataset.}  
Our training dataset is constructed by carefully sampling 3,000 instances from OS-Atlas~\cite{os-atlas}, OS-Genesis~\cite{os-genesis}, GUI-R1~\cite{gui-r1}, and AndroidControl~\cite{android_control}. The dataset spans three major platform categories: Android, Desktop, and Web, thereby ensuring a diverse task distribution and comprehensive coverage of real-world GUI interactions.

\noindent\textbf{Benchmarks and Evaluation Metrics.}  
We evaluate GUI grounding performance on three established benchmarks: ScreenSpot, ScreenSpot-v2, and ScreenSpot-Pro.  
ScreenSpot~\cite{seeclick} contains relatively simple tasks focused on common mobile and desktop interfaces.  
ScreenSpot-v2~\cite{os-atlas} extends this by incorporating more diverse interface layouts and interaction patterns.  
ScreenSpot-Pro~\cite{screenspot-pro} targets professional, high-resolution interfaces that exhibit greater structural and semantic complexity. It is designed to assess model generalization in more realistic GUI environments.
Following the standard evaluation protocol~\cite{seeclick,screenspot-pro}, a prediction is considered correct if the predicted center point falls within the ground-truth bounding box.

\noindent\textbf{Comparison Baselines.}  
We evaluate our model against a wide range of existing methods across different categories: proprietary models (e.g., GPT-4o~\cite{gpt4O}, Claude Computer Use~\cite{anthropic}), general vision-language models (e.g., Qwen2.5-VL~\cite{qwen2.5-vl}), and GUI-specific models with supervised finetuning or reinforcement learning (e.g., OS-Atlas~\cite{os-atlas},  UI-TARS~\cite{ui-tars}, CogAgent~\cite{cogagent}, ShowUI~\cite{showui}, SE-GUI~\cite{enhancing}, GUI-R1~\cite{gui-r1}, InfiGUI-R1~\cite{infigui-r1}, GUI-G1~\cite{gui—g1}).  
All baseline results are collected from official papers or publicly released checkpoints.

\subsection{Experimental Results and Analysis}

\textbf{Main Results}  

We evaluate our model, GUI-Eyes-3B, on three benchmarks—ScreenSpot~\cite{seeclick}, ScreenSpot-v2~\cite{os-atlas}, and ScreenSpot-Pro~\cite{screenspot-pro} to assess its GUI grounding capabilities.
As shown in Table~\ref{tab:screenspot-pro} and Table~\ref{tab:screenspot-full}, GUI-Eyes-3B achieves state-of-the-art performance across all three benchmarks, ranking first in overall accuracy and consistently outperforming prior methods on both text and icon grounding tasks.

On ScreenSpot, it achieves an overall accuracy of 87.8\%, leading across most platform settings.  
On ScreenSpot-v2, GUI-Eyes-3B reaches 88.4\%, showing consistent improvements across all device categories.  
On ScreenSpot-Pro, it demonstrates strong generalization in complex domains, particularly in CAD (48.2\%), development tools (70.8\%), and scientific software (69.4\%).  

Compared to prior RL-based models such as GUI-R1-3B and GUI-G1-3B on the more challenging ScreenSpot-Pro benchmark, GUI-Eyes-3B delivers more balanced and robust performance, particularly in visually cluttered or low-saliency environments.
Overall, these results validate the effectiveness and generalizability of our tool-augmented reasoning framework, underscoring its potential for real-world GUI interaction and automation systems.

\begin{table}[t]
\centering
\begin{tabular}{ccccc}
\toprule
\textbf{Index} & \textbf{$\lambda_{\text{acc}}$} & \textbf{$\lambda_{\text{tool}}$} & \textbf{$\lambda_{\text{format}}$} & \textbf{Accuracy} \\
\midrule
1 & 0.4  & 0.5  & 0.1  & 43.2 \\ 
2 & 0.55 & 0.35 & 0.1  & 44.5 \\ 
3 & \textbf{0.6}  & \textbf{0.3}  & \textbf{0.1}  & \textbf{44.8} \\ 
4 & 0.65 & 0.25 & 0.1  & 42.6 \\ 
5 & 0.7  & 0.2  & 0.1  & 41.2 \\ 
\bottomrule
\end{tabular}
\caption{Grounding accuracy (\%) on ScreenSpot-Pro under different reward-coefficient settings.}
\label{tab:reward_coefficient}
\end{table}

\noindent\textbf{Experimental Analysis}

\begin{table}[t]
\centering
\begin{tabular}{lccc}
\toprule
\textbf{Reward Function} & \textbf{Text Acc} & \textbf{Icon Acc} & \textbf{Overall} \\
\midrule
Center Only &55.7 & 13.9 &\textbf{39.7}        \\
Overlap Only &57.9 & 15.4 &\textbf{41.7}    \\
\textbf{Full (Ours)} & 62.8 & 15.7 & \textbf{44.8} \\
\bottomrule
\end{tabular}
\caption{Grounding accuracy (\%) on ScreenSpot-Pro using different tool reward functions.}\label{tab:reward_ablation}
\end{table}

\begin{figure}[b]
\centering
\includegraphics[width=\linewidth]{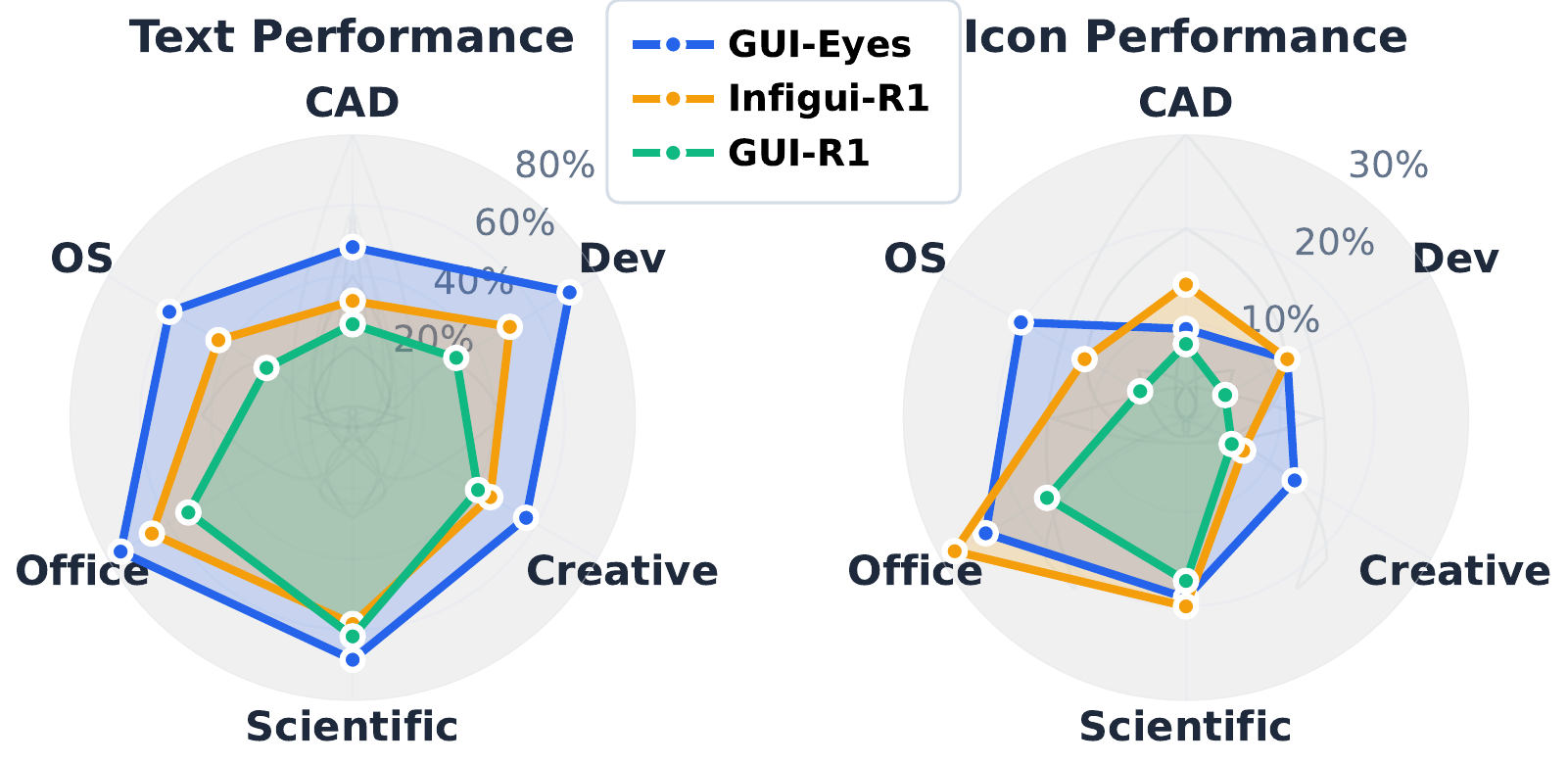}
\caption{ Radar plots comparing the grounding accuracy of GUI-Eyes-3B, Infigui-R1-3B, and GUI-R1-3B on text-based (left) and icon-based (right) queries across domains in the ScreenSpot-Pro benchmark.}
\label{fig:radar}
\end{figure}

\noindent\textbf{Text vs. Icon Performance.}
To further examine the behavior of the model in different query types, we conducted a comparative analysis of grounding performance in text-based versus icon-based queries.
As illustrated in Figure~\ref{fig:radar}, GUI-Eyes-3B achieves substantial improvements in text grounding accuracy across various domains in the ScreenSpot-Pro~\cite{screenspot-pro} benchmark, with particularly notable gains in CAD and development tool scenarios.
Despite being trained on only 3,000 labeled examples, our model surpasses several strong reinforcement learning baselines, including GUI-R1-3B and GUI-G1-3B, the latter trained with 17,000 RL samples, demonstrating strong generalization under limited supervision.

For icon-based queries, GUI-Eyes-3B demonstrates consistent performance gains, although the improvements are slightly smaller than those observed on text tasks. This suggests that the proposed method effectively handles both linguistic and symbolic grounding, leveraging the model’s latent visual understanding to generalize across abstract, icon-driven interface elements. Future efforts could further enhance this capacity by incorporating targeted visual pretraining or lightweight symbol-aware augmentation strategies.

\subsection{Ablation Study}

\noindent\textbf{Ablation Study on Reward Coefficient Sensitivity}

We conduct an ablation study on \textbf{ScreenSpot-Pro} benchmark to examine the sensitivity of the model to different reward coefficients, specifically $\lambda_{\text{acc}}$, $\lambda_{\text{tool}}$, and $\lambda_{\text{format}}$ in Equation~\ref{eq:total_reward}. These parameters control the relative importance of accuracy, tool usage, and format rewards, respectively.

As shown in Table~\ref{tab:reward_coefficient}, the best configuration is obtained with $\lambda_{\text{acc}} = 0.6$, $\lambda_{\text{tool}} = 0.3$, and $\lambda_{\text{format}} = 0.1$, achieving a grounding accuracy of \textbf{44.8\%} on ScreenSpot-Pro. 
Different reward coefficients can influence the model’s performance, particularly with respect to the trade-off between accuracy and tool utilization. Therefore, carefully tuning these weights is essential for achieving optimal results.

\noindent\textbf{Ablation Study on Tool Reward Design}

To better supervise the model's perceptual behavior, we formulate the tool reward $R_{\text{tool}}$ with two key components (see Eq.~\ref{eq:tool_reward}):  
(1) \textit{Center Proximity}, which measures the distance between the selected focus point and the target region;  
(2) \textit{Region Overlap}, which quantifies the spatial intersection between the tool's operation area and the ground-truth bounding box.

We evaluate three reward variants on the ScreenSpot-Pro~\cite{screenspot-pro} benchmark:  
(i) \textbf{Center Only};  
(ii) \textbf{Overlap Only}; and  
(iii) \textbf{Full}, which combines both.

As summarized in Table~\ref{tab:reward_ablation}, using either component in isolation results in limited gains, while the full reward yields significantly better grounding accuracy—especially for text queries.  
These results highlight the importance of jointly guiding both the initial attention point and the tool's coverage region to improve tool use and decision-making effectiveness.

\noindent\textbf{The Impact of Tool Usage in Training}

\begin{figure}[t]
\centering
\includegraphics[width=\linewidth]{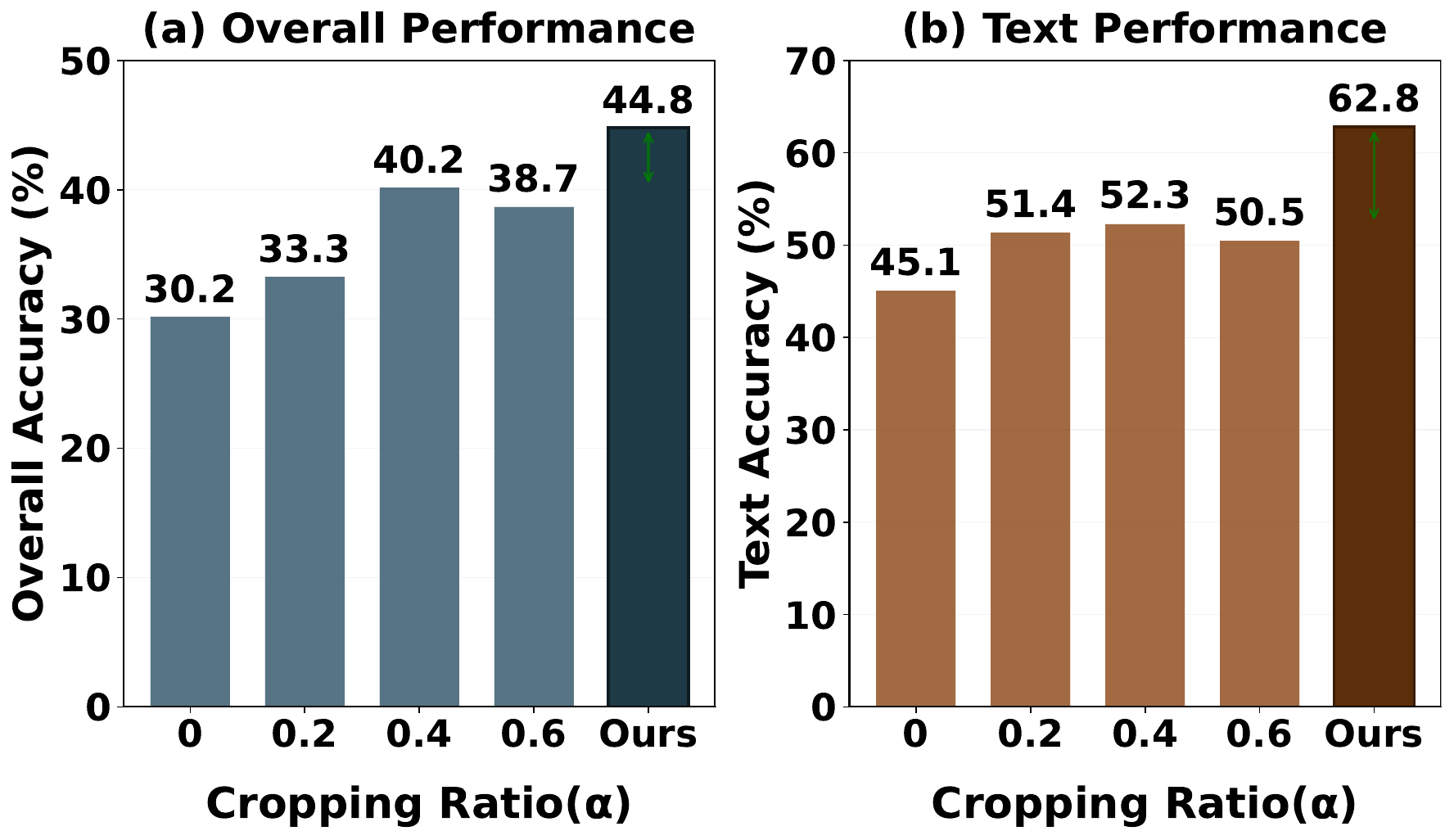}
\caption{Ablation study comparing different tool-usage strategies on ScreenSpot-Pro. 
$\alpha = 0$ denotes no tool usage. 
$\alpha \in \{0.2, 0.4, 0.6\}$ are fixed cropping ratios generated from GUI-R1-3B predictions (static cropping). 
``Ours'' refers to our GUIEyes-3B model, which dynamically learns when and how much to crop.
}
\label{fig:ablation}
\end{figure}

To evaluate the contribution of the tool-based perception mechanism in our framework, we conduct an ablation study by progressively disabling components of the tool learning pipeline. Specifically, we compare the following variants:

\begin{itemize}
  \item \textbf{No Tool Usage:} 
  The agent is restricted from invoking any visual tools during inference and must rely solely on the raw GUI screenshot. This setting corresponds to a cropping ratio of $\alpha = 0$ in Figure~\ref{fig:ablation}, serving as the baseline for evaluating the benefit of tool-based perception.

  \item \textbf{No Tool Training:} Visual tools remain accessible, but the tool invocation policy is no longer learned. Instead, we adopt a fixed heuristic inspired by the DiMo-GUI framework~\cite{dimo}, where the tool input is generated by cropping a region centered on the prediction from a strong pretrained model, GUI-R1-3B. The cropping ratio $\alpha$ is varied to examine the effect of input scale (e.g., $\alpha=0.2, 0.4, 0.6$).
\end{itemize}

We select GUI-R1-3B as a comparison model because it is trained with reinforcement learning on the same data scale (3,000 samples), ensuring fair comparability.
The results are summarized in Figure~\ref{fig:ablation}. As shown, the agent without any tool usage achieves the lowest performance (30.2\% overall accuracy and 45.1\% on text-based queries). When incorporating static cropping based on the pretrained model's prediction, performance improves across different cropping scales, with the best result at $\alpha=0.4$ achieving 40.2\% overall and 52.3\% text accuracy.
In contrast, our method outperforms these static strategies, achieving 44.8\% overall accuracy and 62.8\% on text queries. These findings highlight the importance of learning a dynamic tool policy to support perception refinement and robust decision-making.

\section{Conclusion}

In this work, we propose GUI-Eyes, a reinforcement learning framework that guides multimodal language models to perform structured perception-to-decision reasoning in graphical user interface (GUI) environments. The framework introduces an active perception mechanism, enabling the model to dynamically decide whether to invoke visual tools—such as cropping and zooming—and to configure them adaptively during inference, thereby acquiring more focused and task-relevant observations.
 To support effective tool usage, we design a spatially aware reward function that combines location proximity and region overlap, offering dense and stable optimization feedback.
Extensive experiments demonstrate that GUI-Eyes-3B, trained on only 3,000 labeled samples, achieves 44.8\% accuracy on the ScreenSpot-Pro benchmark, significantly outperforming both supervised and RL-based baselines. These results highlight the framework’s strong generalization ability and data efficiency, underscoring its potential for building scalable and perceptually grounded GUI agents.

\bigskip

\bibliography{aaai2026}

\clearpage
\appendix
\renewcommand{\thefigure}{\arabic{figure}}
\setcounter{figure}{0}
\renewcommand{\thetable}{\arabic{table}}
\setcounter{table}{0}

\section{Appendix}
\subsection{Additional Examples of GUI Tasks}

In this appendix, we present several illustrative examples of different GUI tasks under our active perception framework. These examples demonstrate how the agent strategically invokes visual tools and performs multi-stage reasoning to complete grounding tasks across diverse interface scenarios.

As shown in Figure 1, the agent employs a cropping tool to refine its visual input before making a grounding decision.

As shown in Figure 2, the agent completes the grounding task directly without invoking any visual tools.

As shown in Figure 3, the agent applies a zooming operation to enhance visual clarity before executing the final action.

\begin{figure}[ht]
\centering
\includegraphics[width=\linewidth]{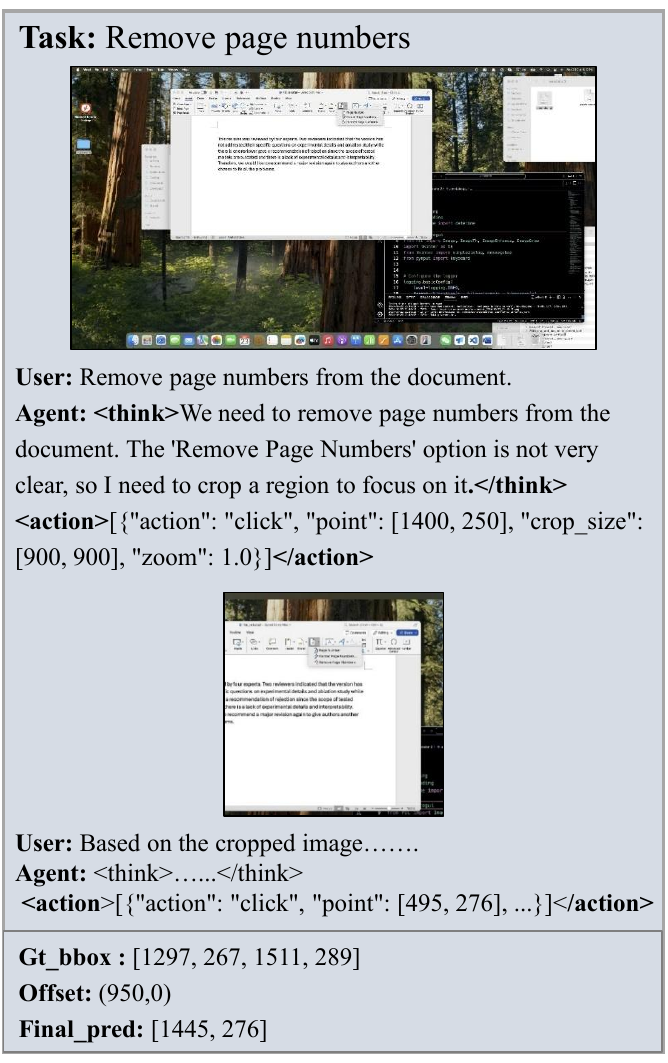}
\caption{An Example of Active Perception with Cropping. Offset maps the point from crop to original image.}
\label{fig:example2}
\end{figure}

\begin{figure}[h]
\centering
\includegraphics[width=\linewidth]{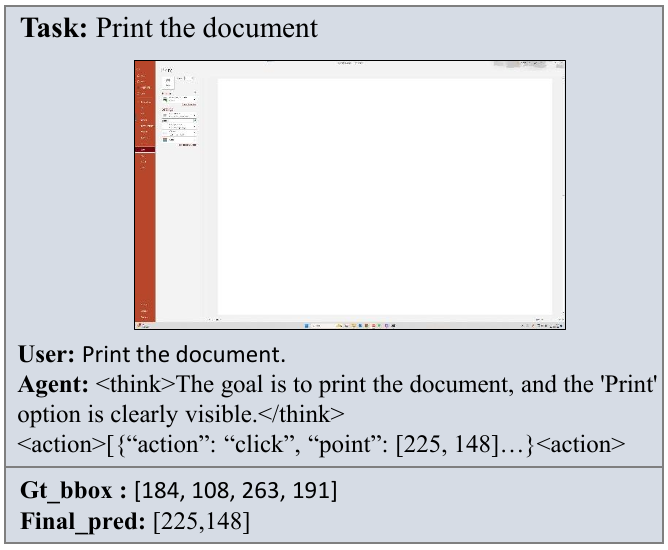}
\caption{An Example of Direct Grounding without Tool.}
\label{fig:example1}
\end{figure}

\begin{figure}[h]
\centering
\includegraphics[width=\linewidth]{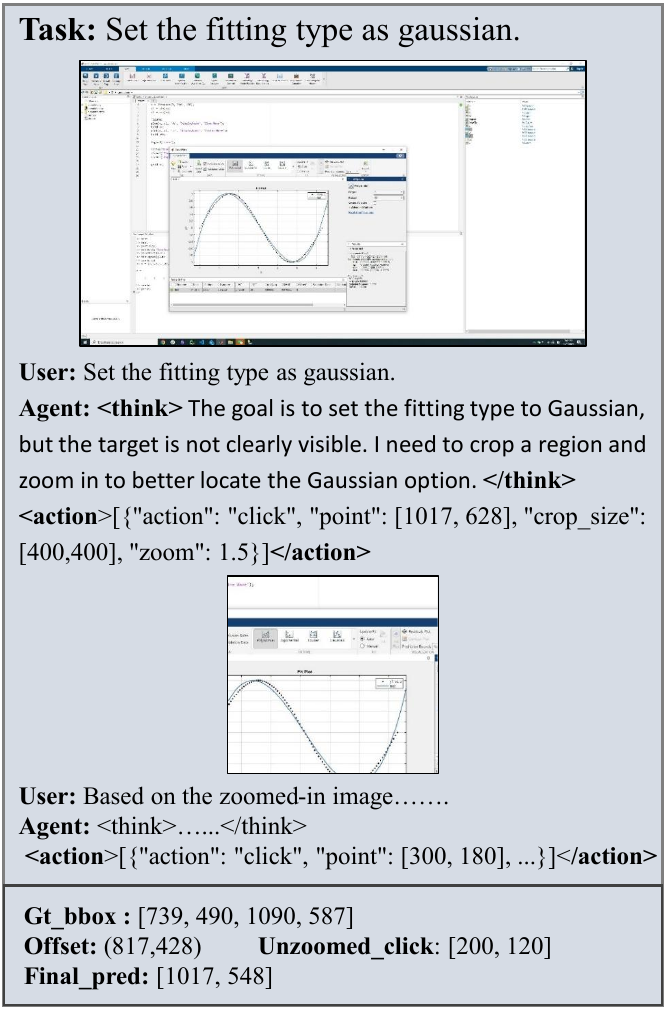}
\caption{An Example of Active Perception with Zooming. Offset maps the point from crop to original image. Unzoomed click refers to the click point adjusted by the inverse of zoom factor.}
\label{fig:example3}
\end{figure}

\newpage
\subsection{Implementation and Training Details}
\vspace{0.5em}

\noindent \textbf{GRPO Hyperparameter Settings}

\vspace{0.5em}
The detailed training configuration for our GRPO-based policy learning is provided in Table~\ref{tab:grpo_hyperparams}. All experiments are conducted using 8 NVIDIA H100-80G GPUs.

\begin{table}[h]
\centering
\small
\begin{tabular}{ll}
\toprule
\textbf{Hyperparameter} & \textbf{Value} \\
\midrule
learning\_rate & $1 \times 10^{-6}$ \\
temperature & 1.0 \\
num\_generations & 6 \\
max\_prompt\_length & 8000 \\
max\_completion\_length & 1000 \\
per\_device\_train\_batch\_size & 1 \\
gradient\_accumulation\_steps & 4 \\
$\epsilon$ (clipping parameter) & 0.2 \\
$\beta$ (KL coefficient) & 0 \\
\bottomrule
\end{tabular}
\caption{GRPO hyperparameter settings used in our training.}
\label{tab:grpo_hyperparams}
\end{table}

%

\noindent \textbf{Reward Coefficient Values}
\vspace{0.5em}

We recall the total reward function defined in Equations~\ref{eq:total_reward} and~\ref{eq:tool_reward}, which combines multiple objectives:

\begin{itemize}
  \item $R_{\text{acc}}$: Task accuracy reward based on the correctness of the final prediction.
  \item $R_{\text{format}}$: Format validity reward to enforce well-formed action outputs.
  \item $R_{\text{tool}}$: Tool usage reward composed of:
  \begin{itemize}
    \item A spatial proximity term measuring distance between the predicted point and the ground-truth box center.
    \item An overlap term measuring the intersection-over-union with the ground-truth box.
  \end{itemize}
\end{itemize}

\noindent The complete reward is computed as:
\begin{equation*}
R(\tau) = \lambda_{\text{acc}} R_{\text{acc}} + \lambda_{\text{format}} R_{\text{format}} + \lambda_{\text{tool}} R_{\text{tool}}
\label{eq:total_reward_1}
\end{equation*}

\begin{equation*}
\begin{aligned}
R_{\text{tool}} =\;& \lambda_{\text{center}} \cdot 
\exp\left( -\alpha \left( \frac{d(c, \text{gt\_bbox})}{\sigma} \right)^2 \right) \\
&+ \lambda_{\text{overlap}} \cdot 
\frac{|\text{crop\_bbox} \cap \text{gt\_bbox}|}{|\text{gt\_bbox}|}
\end{aligned}
\label{eq:tool_reward_2}
\end{equation*}

\noindent
The reward coefficients used in our implementation are as follows:

\begin{equation*}
\begin{aligned}
\lambda_{\text{acc}} &= 0.6 \\
\lambda_{\text{format}} &= 0.1 \\
\lambda_{\text{tool}} &= 0.3 \\
\lambda_{\text{center}} &= 0.7 \\
\lambda_{\text{overlap}} &= 0.3 \\
\alpha &= 1.5 \\
\sigma &= 1.6 \cdot \sqrt{(x_2 - x_1)^2 + (y_2 - y_1)^2}
\end{aligned}
\end{equation*}

Here, $(x_1, y_1)$ and $(x_2, y_2)$ denote the top-left and bottom-right coordinates of the ground-truth bounding box, respectively. The term $\sigma$ represents the scaled diagonal length of the box and serves as the normalization factor for distance-based reward shaping.

All coefficients are selected via grid search on the validation set to ensure stable learning dynamics and generalizable policy behavior.

\vspace{1em}

\subsection{Additional Experimental Results}

Table 1 presents the detailed results of GUI-Eyes-3B on ScreenSpot v1. Despite being trained on only 3K samples, our model achieves the highest overall accuracy (87.8\%) and consistently outperforms prior baselines on both text and icon grounding tasks. As noted in Section 4.2, GUI-Eyes-3B exhibits particularly strong performance on text-based tasks, further demonstrating that our active perception (reinforcement learning) strategy effectively unlocks the underlying capabilities of the base model.

\begin{table}[t]
\centering
\small
\setlength{\tabcolsep}{3.2pt} 
\begin{tabular}{l c c c c c c c}
\toprule
\textbf{Model} & \multicolumn{2}{c}{Mobile} & \multicolumn{2}{c}{Desktop} & \multicolumn{2}{c}{Web} &Overall \\
 & Text & Icon & Text & Icon & Text & Icon \\
\midrule
\multicolumn{5}{l}{\textbf{General Open-source Models}} \\
Qwen2-VL-7B & 61.3 & 39.3 & 52.0 & 45.0 & 33.0 & 21.8 & 42.9 \\
Qwen2.5-VL-3B & - & - & - & - & - & - & 55.5  \\
Qwen2.5-VL-7B & - & - & - & - & - & - & 84.7  \\
\midrule
\multicolumn{5}{l}{\textbf{GUI-Specific Models(SFT)}} \\
CogAgent-18B & 67.0 & 24.0 & 74.2 & 20.0 & 70.4 & 28.6 & 47.4  \\
SeeClick-9.6 & 78.0 & 52.0 & 72.2 & 30.0 & 55.7 & 32.5 & 53.4  \\
UGround-7B & 82.8 & 60.3 & 82.5 & 63.6 & 80.4 & 70.4 & 73.3  \\
ShowUI-2B & 92.3 & 75.5 & 76.3 & 61.1 & 81.7 & 63.6 & 75.1  \\
OS-Atlas-7B & 93.0 & 72.9 & 91.8 & 62.9 & 90.9 & 74.3 & 82.5  \\
Aguvis-7B & 95.6 & 77.7 & 93.8 & 67.1 & 88.3 & 75.2 & 84.4  \\
 UI-TARS-2B & 93.0 & 75.5 & 90.7 & 68.6 & 84.3 & 74.8 & 82.3 \\
\midrule
\multicolumn{5}{l}{\textbf{GUI-Specific Models(RL)}} \\
UI-R1-3B & 95.6 & \textbf{84.7} & 90.2 & 59.3 & 85.2 & 73.3 & 83.3  \\
GUI-R1-3B & - & - & 93.8 & 64.8 & 89.6 & 72.1 & -  \\
GUI-R1-7B & - & - & 91.8 & 73.6 & 91.3 & 75.7 & -\\
\midrule
\textbf{GUIEyes-3B} & \textbf{96.7} & 81.7 & \textbf{96.4} & \textbf{77.1}	& \textbf{92.4}	& \textbf{77} & \textbf{87.8}  \\
\bottomrule
\end{tabular}
\caption{Performance comparison details on ScreenSpot. Bold highlights the best results.}
\label{tab:screenspot}
\end{table}

\end{document}